\documentclass[conference]{IEEEtran}
\IEEEoverridecommandlockouts
\usepackage{cite}
\usepackage{amsmath,amssymb,amsfonts}
\usepackage{algorithmic}
\usepackage{graphicx}
\usepackage{textcomp}
\usepackage{xcolor}
\usepackage[ruled,vlined]{algorithm2e}
\usepackage{subfigure}
\usepackage{hyperref}
\usepackage{flushend}
\usepackage{booktabs}

\def\BibTeX{{\rm B\kern-.05em{\sc i\kern-.025em b}\kern-.08em
    T\kern-.1667em\lower.7ex\hbox{E}\kern-.125emX}}
\begin{document}

\title{Autonomous Docking Method via Non-linear Model Predictive Control}

\author{
\IEEEauthorblockN{Roni Permana Saputra}
\IEEEauthorblockA{\textit{Research Center for Smart Mechatronics}\\
\textit{National Research and Innovation Agency}\\
Bandung, Indonesia\\
Email: roni.permana.saputra@brin.go.id\\
 ORCID: \href{https://orcid.org/0000-0001-6989-8830}{https://orcid.org/0000-0001-6989-8830}}
\\   
\IEEEauthorblockN{Eko Joni Pristianto}
\IEEEauthorblockA{\textit{Research Center for Telecommunication}\\
\textit{National Research and Innovation Agency}\\
Bandung, Indonesia\\
Email: eko.joni.pristianto@brin.go.id}
\and
\IEEEauthorblockN{Midriem Mirdanies}
\IEEEauthorblockA{\textit{Research Center for Smart Mechatronics}\\
\textit{National Research and Innovation Agency}\\
Bandung, Indonesia\\
Email: midriem.mirdanies@brin.go.id}\\ 
\\
\IEEEauthorblockN{Dayat Kurniawan}
\IEEEauthorblockA{\textit{Research Center for Telecommunication}\\
\textit{National Research and Innovation Agency}\\
Bandung, Indonesia\\
Email: dayat.kurniawan@brin.go.id}
}

\maketitle

\begin{abstract}
This paper presents a proposed method of autonomous control for docking tasks of a single-seat personal mobility vehicle.
We proposed a non-linear model predictive control (NMPC) based visual servoing to achieves the desired autonomous docking task.
The proposed method is implemented on a four-wheel electric wheelchair platform, with two independent rear driving wheels and two front castor wheels.
The NMPC-based visual servoing technique leverages the information extracted from a visual sensor as a real-time feedback for the NMPC to control the motion of the vehicle achieving the desired autonomous docking task.
To evaluate the performance of the proposed controller method, a number of experiments both in simulation and in the actual setting.
The controller performance is then evaluated based on the controller design requirement.
The simulation results on autonomous docking experiments show that the proposed controller has been successfully achieve the desired controller design requirement to generate realtime trajectory for the vehicle performing autonomous docking tasks in several different scenarios.
\end{abstract}


\section{Introduction}
\label{sec:intro}

The use of environmentally friendly modes of transportation, especially vehicles with power electric-based drives are currently starting to be actively encouraged.
One type of electric vehicle that gaining popularity is the personal mobility vehicles (PMVs). 
This vehicle type consists of one or more wheels which driven by electric motor that travels around 2 m/s---very close to normal human walking speed. 
This vehicle is commonly used as a mode of transportation for short-distance trips in urban environments or in a specific and limited area.
This type of vehicle provides various advantages from an environmental standpoint to socio-economic benefits.

\begin{figure}[h]
  \centering
  \includegraphics[width=0.75\linewidth]{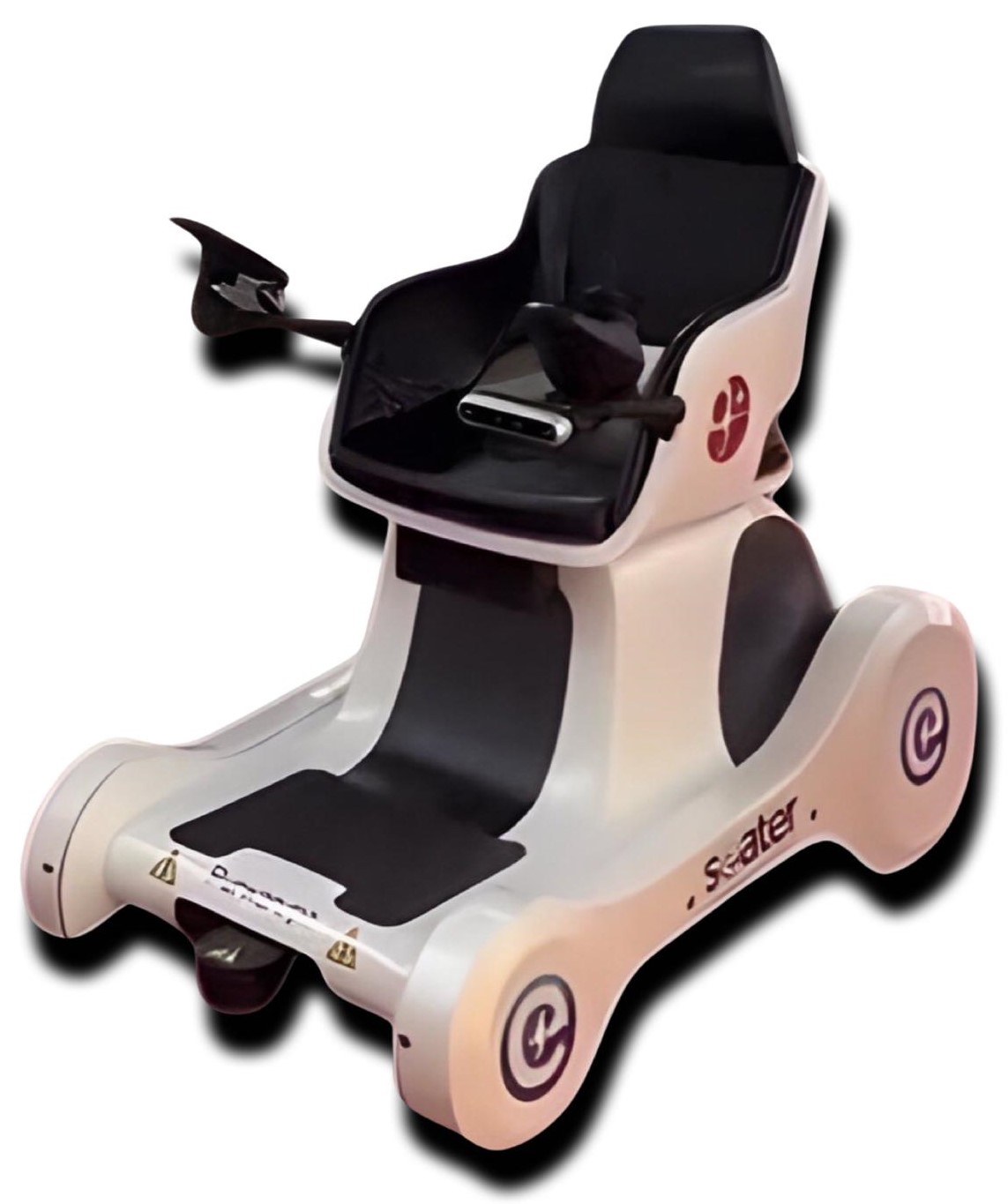}
  \caption{A single passenger personal mobility vehicle platform---with two independent rear driving wheels and two front castor wheels---equipped with an onboard camera sensor for visual servoing feedback.}
\vspace{-1.5em}
\end{figure}

Over the past few decades, technological advancements in control, sensing, and artificial intelligence, particularly those applied to transportation, have encouraged the development of autonomous vehicle technology. 
The effect of this technology is expected to benefits us through the improvement of transportation safety, increasing driving efficiency and transportation convenience, reducing congestion, and other various benefits.
One of the important features required for an autonomous vehicle is the capability of the vehicle to navigate and dock itself autonomously to a charging dock station.
The vehicles can be recharged by planning and arranging effectively to  maximize  its  working  efficiency.
This paper adresses the autonomous docking problem for an autonomous personal mobility vehicle.
We proposed a non-linear model predictive control (NMPC) based visual servoing to achieves the desired autonomous docking task.
The NMPC-based visual servoing technique leverages the information extracted from a visual sensor as a real-time feedback for the NMPC to control the motion of the vehicle achieving the desired autonomous docking task.
In forthcoming development, the proposed approach is expected to be fully implemented in a four-wheeled personal mobility vehicle. 
This vehicle consists of two independent rear drive wheels and two front castor wheels. Additionally, the vehicle is equipped with an RGB-D camera sensor, which provides visual servoing input as seen in Fig.~1.

\section{Related Work}
\label{sec:related-work}

A number of studies have been conducted in the field of optimal control methods due to their demonstrated success in various applications.
The methods are commonly utilized, particularly in situations where the dynamics of the controlled systems are complex or when there are numerous operating constraints that need to be achieved~\cite{rigatos2017,wen2019,Zanon2017,saputra2021hierarchical}.

Model predictive control (MPC) has become increasingly popular and is now widely utilized across various applications~\cite{rossiter2017model}.
MPC achieves the desired system behavior by utilizing a system model to forecast future states over a finite time horizon. It then optimizes control decisions based on a specified constrained cost function~\cite{rawlings2009model}.
A key advantage of MPC is its ability to incorporate multiple constraints into the controller formulation. This enables the controller to effectively handle complex problems, including intricate desired behavior, environmental constraints, and hardware limitations.
Numerous studies have utilized different forms of MPC for controlling non-holonomic mobile robots. These studies have explored various applications, including point stabilization, path and trajectory tracking, and collision avoidance~\cite{gu2005stabilizing,huynh2017comparative,li2015trajectory,yu2018mpc,hirose2018mpc,7487274,ribeiro2019nonlinear, fukushima2013model}.

Neunert et al. in~\cite{7487274} introduced a framework in their work that addresses the challenge of real-time nonlinear model predictive control (MPC) for mobile robots. Their approach efficiently solves both trajectory optimization and tracking problems in a unified manner.
The framework presents an iterative optimal control algorithm known as Sequential Linear Quadratic (SLQ) to address the nonlinear Model Predictive Control (MPC) problem.
In their study, Li et al. suggest employing a primal-dual neural network to address the Quadratic Programming (QP) problem within a finite receding horizon. This approach aims to enable trajectory control of a mobile robot system~\cite{li2015trajectory}.
The proposed approach aims to ensure convergence of the formulated constrained quadratic programming (QP) cost function to the precise optimal values. It also demonstrates effective performance on a practical mobile robot system.
Hirose et al. also introduce a work that integrates neural networks into the model predictive control (MPC) framework~\cite{hirose2018mpc}.
The authors in~\cite{hirose2018mpc} employs a deep neural network to acquire the MPC policy, which differs from the approach taken in~\cite{li2015trajectory}. 

With the progression of technology, there is a corresponding enhancement in the capabilities and processing speed of camera sensors and computers. 
The progression of technology has resulted in the development of advanced image-processing algorithms capable of extracting and interpreting characteristics within unprocessed visual data obtained from camera sensors. 
One of the techniques employed is marker detection and posture estimation with the ArUco marker.
The term "ArUco marker" denotes a quadrilateral marker that has a wide black border and an inside binary grid, which serves to indicate its distinct identification (id) as proposed by Garrido-Jurado et al in~\cite{GARRIDOJURADO20142280}.
The Aruco marker has been shown to be capable of producing accurate pose estimation~\cite{wubben2019accurate,gonccalves2020precise} and is effective in detecting several markers simultaneously~\cite{romero2018speeded}.
ArUco markers have been utilized in many studies, such as the work conducted by Miranda et al., where they employed ArUco markers during the landing phase of an Autonomous Navigation System for a Delivery Drone~\cite{miranda2022autonomous}. 
Furthermore, Volden et al. have successfully incorporated ArUco markers into the autonomous docking system of unmanned surface vehicles (USVs)~\cite{volden2022vision}.

\section{Problem Formulation}
\label{sec:problem-definition}

In this section, we derived the kinematic model of the vehicle platform used as the predictive model of the controller.

\begin{figure}[t]
  \centering
  \includegraphics[width=0.65\linewidth]{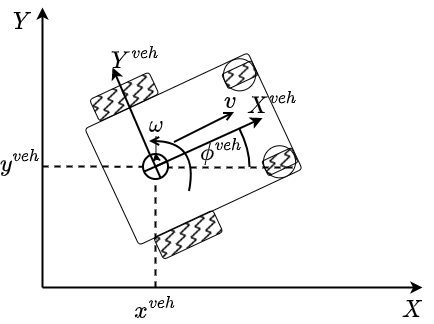}
  \caption{The kinematics model of the SEATER platform approximated as a simple unicycle model.}
  \label{fig:seater-unicycle-model}
\vspace{-1.5em}
\end{figure}

\subsection{Vehicle Kinematic model}

SEATER is a non-holonomic  wheeled vehicle with differential drive configuration (see Fig.~\ref{fig:seater-unicycle-model}).
To model the kinematic of SEATER, we use simple unicycle model, in which the state vector of the vehicle is given by:
\begin{equation} 
\textbf{x}^{veh}(t)=
\begin{bmatrix}
x^{veh}(t)\\ 
y^{veh}(t)\\ 
\phi^{veh}
(t)
\end{bmatrix}
\end{equation}
The first two (spatial) components of the state 
$\textbf{x} = [x, y, \phi]^{T}$  represent the position of the vehicle in the plane while the angle $\phi$ corresponds to the heading of the vehicle.
The vehicle is controlled with input:  
\begin{equation} 
    \textbf{u}(t)=
    \begin{bmatrix}
    v(t)\\ 
    \omega(t)
    \end{bmatrix}
\end{equation}
where $v$ and $\omega$ are forward velocity and angular velocity of the vehicle, respectively.

The vehicle dynamics are derived as:

\begin{equation}\label{eq:dynamics}
    \dot{\mathbf{x}}^{veh}(t) = \mathbf{f}\big(\mathbf{x}^{veh}(t),\, \mathbf{u}(t) \big) =
    \begin{bmatrix}
    v(t)\cos \phi^{veh}(t) \\ 
    v(t)\sin \phi^{veh}(t) \\ 
    \omega(t) 
    \end{bmatrix},
\end{equation}

or, in discrete time:

\begin{equation}
\begin{split}
    \mathbf{x}^{veh}_{|k+1} = &\,\,\, \mathbf{f}(\mathbf{x}^{veh}_{|k},\mathbf{u}_{|k})
    \\
    = &
    \underbrace{\mathbf{x}^{veh}_{|k}}_\textup{current state}
    +
    \Delta t
    \underbrace{\begin{bmatrix}
            v_{|k}\cos\phi^{veh}_{|k}\\ 
            v_{|k}\sin\phi^{veh}_{|k}\\ 
            \omega_{|k}
        \end{bmatrix}}_\textup{state transition}.
\end{split}
\label{eq:discrete-dynamic-model}
\end{equation}

\begin{figure}[t]
  \centering
  \includegraphics[width=0.75\linewidth]{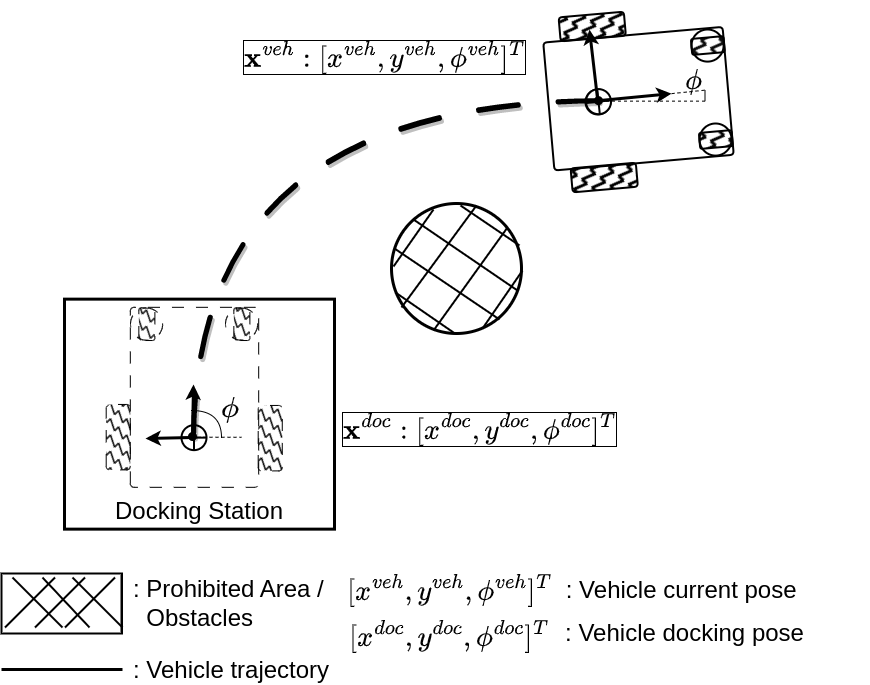}
  \caption{Autonomous docking problem formulation. This study is focused on the proposed method for controlling a personal mobility vehicle platform to autonomously perform a docking operation to a designated station.}
  \label{fig:problem-formulation}
\vspace{-1.5em}
\end{figure}

\subsection{Autonomous Docking as A Nonlinear Model Predictive Control Problem}
\label{subsec:casualty-approach-problem}
The autonomous docking problem --- illustrated in Fig.~\ref{fig:problem-formulation} --- can be formulated as an optimal control problem, where the high-level objective is to minimise the state deviation from the vehicle's current pose, $\mathbf{x}^{veh}$, to the vehicle's target docking pose, $\mathbf{x}^{doc}$:

\begin{equation}\label{eq:overall_objective}
    J = ||\mathbf{x}^{veh} - \mathbf{x}^{doc}||^2
\end{equation} 

To ensure safety during the autonomous docking, 
the vehicle needs to avoid colliding with all possible obstacles (i.e. prohibited area in Fig. 3) in all directions, so that:
\begin{equation}
    \mathbf{x}^{veh}_{|k}
    \in
    {X}^{free}
\end{equation}
where ${X}^{free}$ is the set by all possible vehicle states in which the vehicle is free from collision.

The autonomous docking problem can therefore be written as a model predictive control problem as follows:

\begin{equation}
    \begin{split}
        \underset{\mathbf{u}}{\min} \; \; J_N(\mathbf{x}_0,\mathbf{u}) & = \sum_{k=0}^{N-1} \left \| \mathbf{x}^{veh}_{|k}-\mathbf{x}^{doc} \right \|_{\mathbf{Q}}^{2} + \left \| \mathbf{u}_{|k}-\mathbf{u}^{t} \right \|_{\mathbf{R}}^{2}\\
        \textup{s.t.} \;  \;  \mathbf{x}^{veh}_{|k+1} & =\mathbf{f}(\mathbf{x}^{veh}_{|k},\mathbf{u}_{|k}),  \;  \; \forall k \in [0,N-1]\\
                      \mathbf{x}^{veh}_{|0}     & =\mathbf{x}_0\\
                      \mathbf{u}_{|k}       & \in U,  \;  \;  \forall k \in [0,N-1]\\
                      \mathbf{x}^{veh}_{|k}     & \in X^{free},  \;  \;  \forall k \in [0,N]\\
    \end{split}
\label{eq:OCP-problem}
\end{equation}
where $\left \| \mathbf{x}^{veh}_{|k}-\mathbf{x}^{doc} \right \|_{\mathbf{Q}}^{2}$ and $\left \| \mathbf{u}_{|k}-\mathbf{u}^{t} \right \|_{\mathbf{R}}^{2}$ are the functions of the state deviation and the control effort, respectively.
The expression $ \left \| \mathbf{A} \right \|_{\mathbf{B}}^{2} \equiv  \mathbf{A}^{T}\mathbf{B}\mathbf{A}$.
The matrices $\mathbf{Q}$, $\mathbf{R}$, and $\mathbf{P}$ are positive definite symmetric weighting-matrices of the appropriate dimensions. The weighting matrices are acquired through the heuristic process of  hyperparameters tuning.

\subsection{Realtime Control Feedback via Position Based Visual Servoing (PBVS)}
\label{subsec:visual-servoing-control-feedback}
In order to provide real time feedback to the NMPC controller---i.e., the vehicle's current pose, $\mathbf{x}^{veh}$, to the vehicle's target docking pose, $\mathbf{x}^{doc}$---position based visual servoing (PBVS) is utilised in this work.
In PBVS method, the visual feedback or image measurements are used to determine the target pose of the target with respect to the camera frame which have a fixed transformation to the vehicle frame.
The error between the current and the target pose is defined in the task space of the vehicle.
Hence, the error is a function of pose parameters:
\begin{equation}
    \mathbf{e}(t) = \mathbf{x}(t)-\mathbf{x}^{doc}
\end{equation}

\begin{figure}[t]
  \centering
  \includegraphics[width=0.8\linewidth]{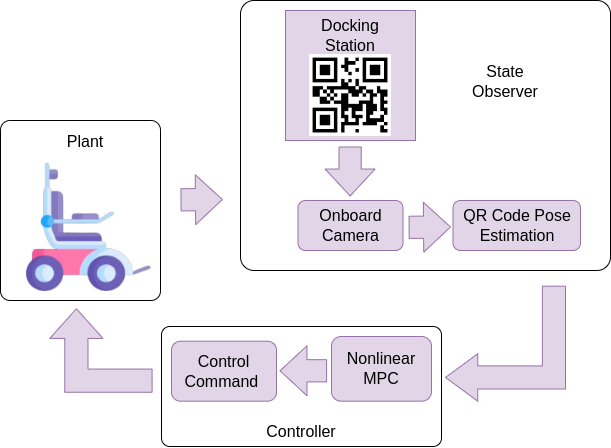}
  \caption{System architecture of the proposed NMPC-based visual servoing control for autonomous docking task.}
  \label{fig:system-architecture}
\vspace{-1.5em}
\end{figure}

\section{Proposed Method}
\label{sec:proposed-method}
\subsection{System Architecture}
We propose a marker based visual servoing with nonlinear model predictive control.
Fig.~\ref{fig:system-architecture} illustrates the system architecture for
 the proposed NMPC-based visual servoing control for autonomous docking task.
In general, the system used in this work consists of two main components, (i) Marker Detector and Pose Estimation, and (ii) Nonlinear Model Predictive Controller.

\subsection{Marker Detector and Pose Estimation}
In this study, we utilise an Intel RealSense D455 camera  to detect an ArUco marker and determine its estimated position relative to the camera frame. 
This position provides a visual reference for the autonomous docking process.
The application program discussed in this paper was developed using the Python programming language and the OpenCV library. 
Additionally, we conducted a camera calibration procedure to compensate for lens errors. 
This procedure involved using a chessboard image captured from multiple angles using the Intel RealSense D455 camera.

\begin{figure}[t]
  \centering
  \includegraphics[width=0.5\linewidth]{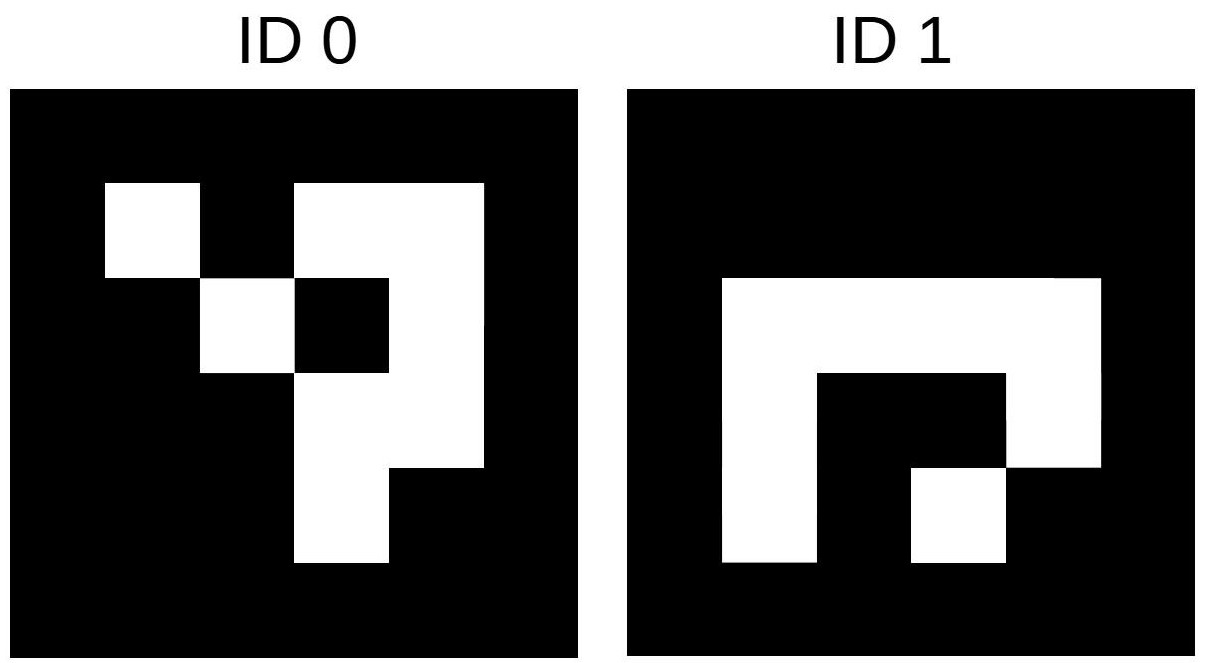}
  \caption{Two examples of generated ArUco markers used for position estimation based on visual features.}
  \label{fig:ArUco-marker}
\vspace{-1.5em}
\end{figure}

In order to validate the position estimation of the ArUco marker, we utilized a marker detection program to detect the ArUco marker, as shown in Fig.~\ref{fig:ArUco-marker}.
The marker was previously printed and its actual length was measured in centimeters. 
Subsequently, the camera's distance from the marker can be determined by calculating its xyz coordinates in centimeters. 
In the event that multiple markers are detected, the program will prioritize the marker with the lowest identification number. 
The xyz coordinates will be leveraged as a reference for guiding the vehicle toward the marker.

\subsection{Nonlinear Model Predictive Algorithm} 
The discrete NMPC controller here is defined as an OCP with a finite control horizon $N$, which evaluates the vehicle state $\mathbf{x}^{veh}$ at every sampling instant $k$.
Then, the optimal control, $\mathbf{u}^*$, for the vehicle is produced at every time step by solving the OCP with respect to the decomposed-objective function at this respective stage, $J_{N}^{s}$, which satisfies the optimal value, $V_{N}^{s}(\hat{\mathbf{x}})$ (i.e. minimising the output of the cost function, $J_{N}^{s}$, s.t. constraints). 
The first element of the produced optimal control trajectory, $\mathbf{u}^*_0$, is then applied to the system. 
Algorithm 1 summarises the proposed approach.

\begin{algorithm}[t]
\caption{Nonlinear Model Predictive Control}
\begin{algorithmic}
    \STATE \textbf{MPCInit:} \newline 
    \hspace*{1em} ControlHorizon := $N$ \newline 
    \hspace*{1em} Define the initial vehicle state: \newline
        \hspace*{1em} \hspace*{1em} ${\mathbf{x}}^{veh}(0):=x_0 \in X^{n_x}$  \newline 
    \hspace*{1em} Get the target state:  \newline 
        \hspace*{1em} \hspace*{1em} $\hat{\mathbf{x}}^{doc}(0) \leftarrow $ StateObserver \newline 
    \hspace*{1em} Define the initial control: $\mathbf{u}_0$  \newline 
    \hspace*{1em} Apply $\mathbf{u}_0$ to the system. \newline
    \textbf{for} {{every sampling instant $k = 1, 2,...$}} \textbf{do} \newline
    \hspace*{1em} Estimate the states $\mathbf{x}^{veh}(k)$ and $\mathbf{x}^{doc}(k)$:  \newline 
    \hspace*{1em} \hspace*{1em} $[\hat{\mathbf{x}}^{veh}(k), \hat{\mathbf{x}}^{doc}(k)]\leftarrow $ StateObserver  \newline 
    \hspace*{1em} \textbf{Solve OCP:} \newline
    \hspace*{1em} Find the optimal control horizon \newline
    \hspace*{3em} $\mathbf{u}^{*} = \{\mathbf{u}^{*}_0, \cdots , \mathbf{u}^{*}_{N-1}\}\in U^{N}$,
    \newline
    \hspace*{1em} which satisfies
    \newline
    \hspace*{3em}  $J_{N}(\hat{\mathbf{x}},\mathbf{u}^{*}) = V_{N}(\hat{\mathbf{x}})$.
    \newline
    \hspace*{3em}
    \textup{s.t.} constraints
    \newline
    \hspace*{1em} Apply $\mathbf{u}^{*}_0$ to the system.
\end{algorithmic}
\label{alg:HiDO-MPC}
\end{algorithm}

\subsection{Controller Design Requirement and Objective Function} 
The autonomous docking task 
There are two main design requirements that the designed controller needs to achieve the autonomous docking task succesfully, position accuracy and heading accuracy.
We introduce two specific cost functions to the desired behaviours that correspond to the controller design requirements:

\begin{itemize}

    \item \textbf{Distance-to-point objective}: Given the current vehicle position ${\mathbf{p}^{veh}}=[x^{veh},y^{veh}]^\intercal$ and the target vehicle docking position ${\mathbf{p}^{doc}}=[x^{doc},y^{doc}]^\intercal$, minimising the distance from the current vehicle position to the target position will drive the vehicle to approach final target position for docking operation.
    The distance between these two points in 2D space can be defined as the Euclidean norm:
    \begin{equation}
      F_1= \left \| \mathbf{p}^{veh}-\mathbf{p}^{doc} \right \|^{2}
    \label{eq:point-obj}
    \end{equation}
    
    \item \textbf{Heading objective}: Given the current vehicle orientation $\phi^{veh}$ and the target docking orientation $\phi^{doc}$, minimising the difference between these orientations $\Delta$ ensures that the vehicle could perform the docking operation in the feasible heading:
    \begin{equation}
        F_2 = \Delta(\phi^{veh},\phi^{doc})=\pi-\left | \left | \phi^{veh}-\phi^{doc} \right | -\pi \right |
    \label{eq:heading-obj},
    \end{equation}
    where:
    \begin{equation*}
        \phi \in [ -\pi, \pi].
    \end{equation*}
    so that it takes into account the periodicity of angles (i.e. angle wrapping).

\end{itemize}

\subsection{Solving the Optimal Control Problem (OCP) via Non-Linear Programming using CasADi}
\label{sec:implementation-HiDO-MPC-as-NLP}
In this study, we use $\texttt{CasADi}$ API~\cite{Andersson2018} to compute the real-time optimisation problem in order to solve the finite OCP problem in the proposed NMPC. The $\texttt{ipopt}$ solver is used by the API to compute the OCP.
The OCP problem is formulated as a standard non-linear programming (NLP). We recommend the reader to~\cite{saputra2021hierarchical} and~\cite{Andersson2018} for a more detailed formulation. 

\section{Experimental Setup}
\label{sec:experimental-setup}
In order to evaluate the efficacy of the proposed method, a series of experimental scenarios were set up and executed on the SEATER platform. 
Initially, the performance of the proposed marker-based pose estimation is assessed.
Furthermore, an assessment is conducted to evaluate the performance of the proposed docking controller in accordance to the designated requirements for controller design.

To evaluate the controller performance, we designed several experiment scenarios to emphasize each of the design requirement points. 
\begin{enumerate}
    \item \textbf{Position Deviation (Shifting):}
    The controller is required to generate realtime trajectory for the vehicle from initial pose to achieve the desired docking pose within the position deviation tolerance. 
    \item \textbf{Heading Deviation (Tilting):}
    The controller is required to generate realtime trajectory for the vehicle from initial pose to achieve the desired docking heading angle within the heading angle deviation tolerance.
\end{enumerate}

\section{Results and Discussion}
\label{sec:proposed-method}

\subsection{Marker-based Visual Pose Estimation}
\label{sec:proposed-method}

\begin{figure}[t]
  \centering
  \subfigure[]{
  \includegraphics[width=0.4\linewidth]{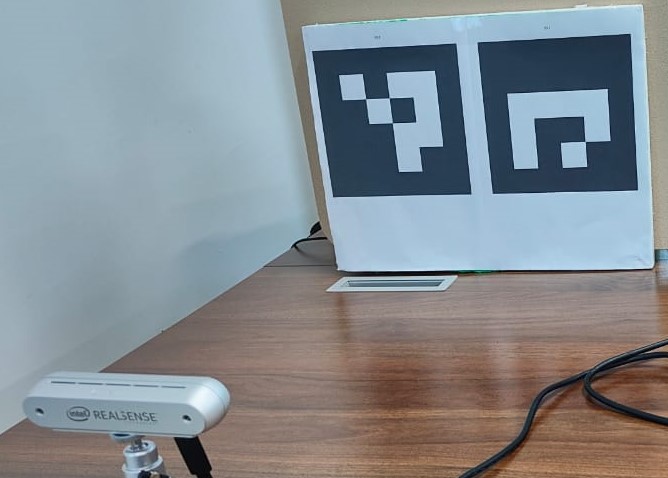}
  }
  \subfigure[]{
  \includegraphics[width=0.4\linewidth]{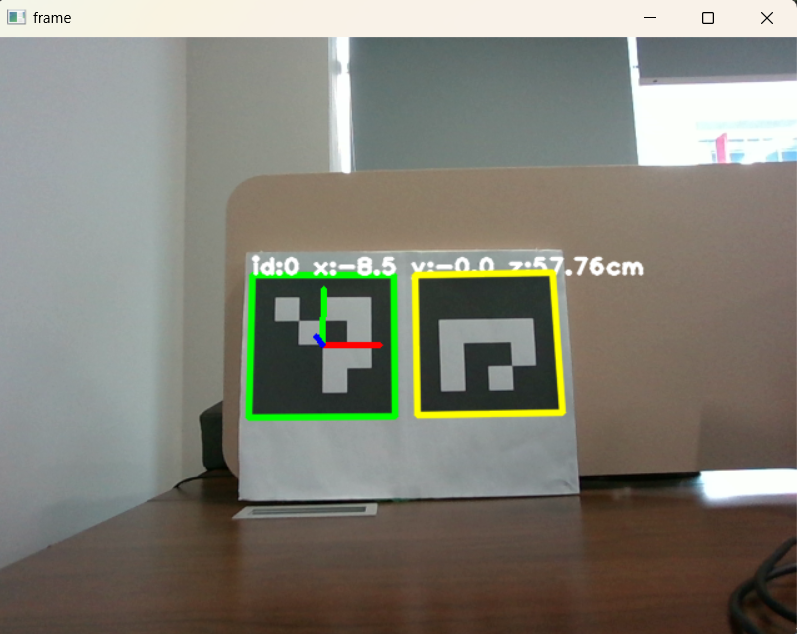}
  }
  \vspace{-0.5em}
  \caption{ArUco marker detection experiment (a) camera position to marker; (b) results of marker detection and pose estimation.}
  \label{fig:ArUco-marker-detection-experiment}
\vspace{-1.5em}
\end{figure}

Software has been develop to detect the ArUco marker and calculate pose estimation which produces xyz coordinates in cm as seen in Fig.~\ref{fig:ArUco-marker-detection-experiment}.
Experiments have been carried out to detect ArUco markers as shown in Fig.~\ref{fig:ArUco-marker-detection-experiment}.a which shows that the system can detect well even though the distance between the camera and marker is sufficient Far. 
The result of the pose estimation calculation can be seen in Fig.~\ref{fig:ArUco-marker-detection-experiment}.b, which is shown in x coordinates, y coordinates and z coordinates or the distance between the marker and the camera in cm.

\begin{figure}[t]
  \centering
  \subfigure[]{
  \includegraphics[width=0.46\linewidth]{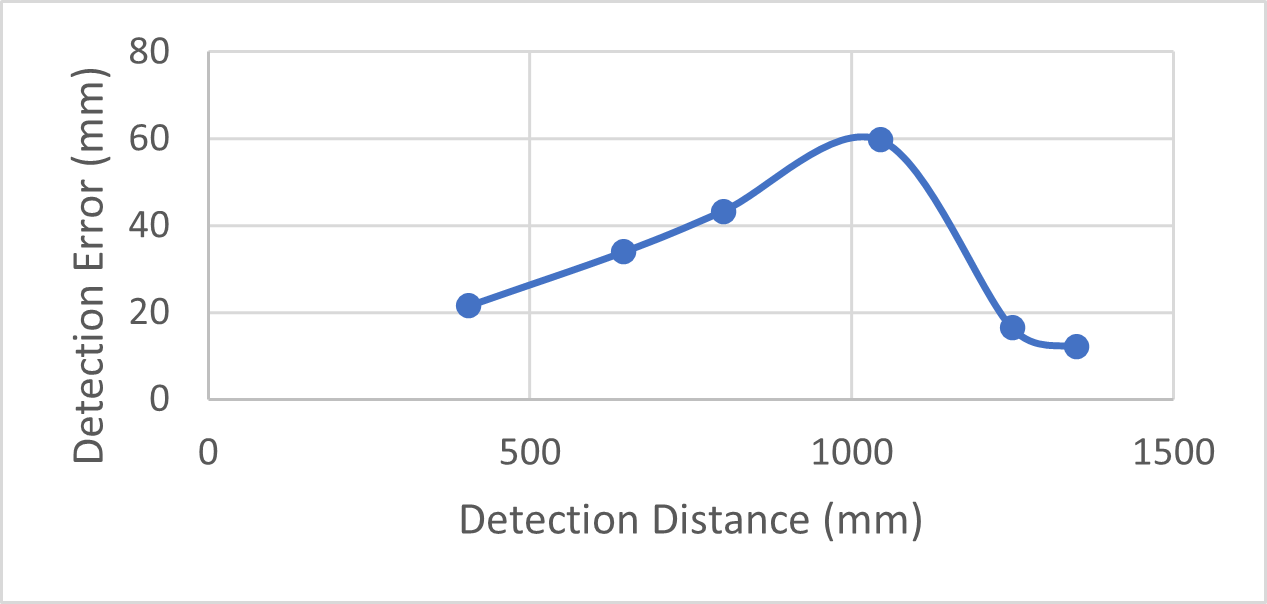}
  }
  \subfigure[]{
  \includegraphics[width=0.46\linewidth]{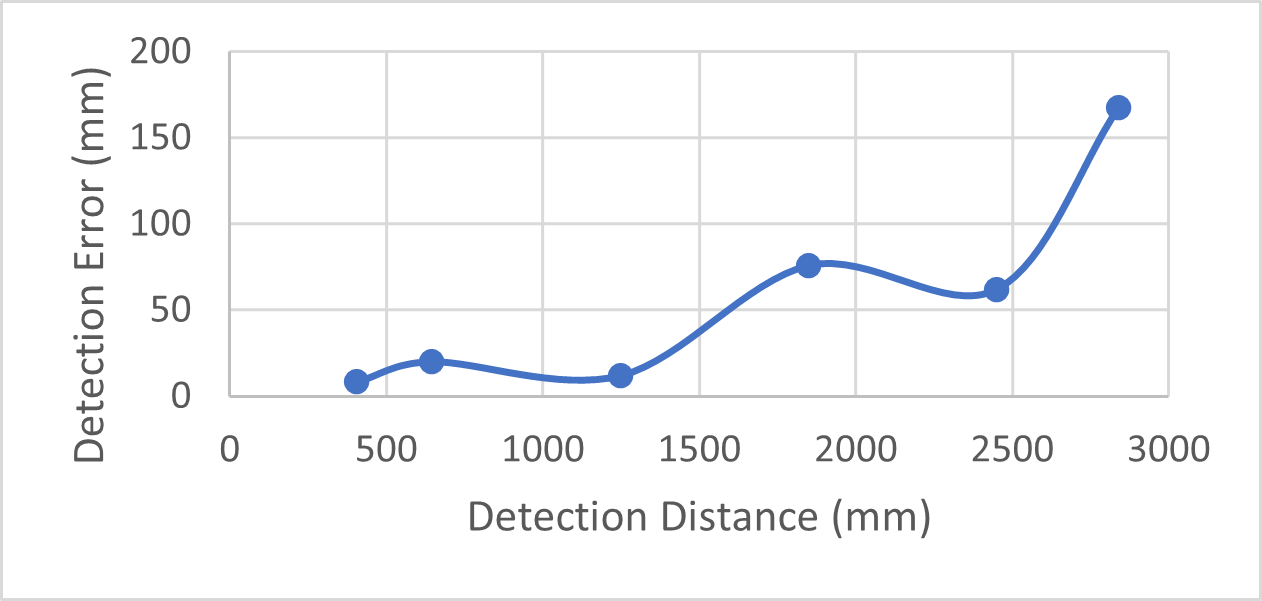}
  }
  \subfigure[]{
  \includegraphics[width=0.46\linewidth]{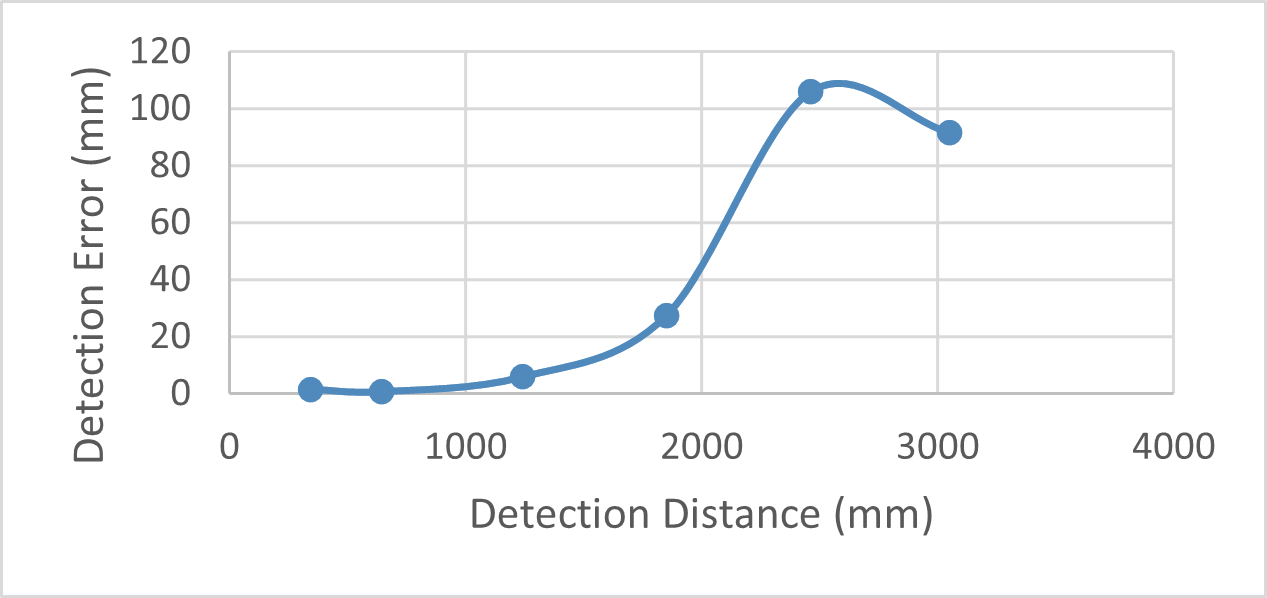}
  }
  \vspace{-0.5em}
  \caption{The relationship between detection distance and detection inaccuracy; (a) Detection inaccuracy for marker with size 50 mm; (b) Detection inaccuracy for marker with size 100 mm; (c) Detection inaccuracy for marker with size 150 mm. }
  \label{fig:ArUco-marker-detection-error}
\vspace{-1.5em}
\end{figure}

We investigated the correlation between marker size and the Intel RealSense D455 camera's minimum and maximum detection distance.
This assessment could be used to identify the optimal marker size for the autonomous docking application.
Table~\ref{tab:aruco-experiments} provides the findings of the evaluation of the relationship between marker size and minimum and maximum detection distance.
Additionally, the relationship between detection distance and detection inaccuracy was investigated as well.
Fig~\ref{fig:ArUco-marker-detection-error} shows the evaluation results of the correlation between detection distance and detection error for three different marker sizes.
Based on the conducted experiments, it has been shown that using marker size dimensions of 100 x 100 mm results in a more consistent detection for the purpose of pose estimation. 
Specifically, the detection accuracy is found to be higher when the marker is located at distances ranging from 350 mm to 1250 mm, as this range yields a reduced margin of error in the estimated pose.
Hence, the obtained results may be used as a valuable point of reference for determining the appropriate marker size and optimal operational distance in the context of the proposed autonomous docking technique.

\begin{table}[]
\centering
\caption{The relationship between marker size and maximum detection distance.}
\label{tab:aruco-experiments}
\resizebox{\linewidth}{!}{%
\begin{tabular}{@{}ccc@{}}
\toprule
\parbox{0.3\linewidth}{
Marker Size \\ (mm x mm)
}
& 
\parbox{0.3\linewidth}{
Min. Detection \\Distance (mm) 
}
& 
\parbox{0.3\linewidth}{
Max. Detection \\Distance (mm)
}
\\ \midrule
50 x 50           &   50  &  1500                     \\
100 x 100           &  90   &   2840                    \\
150 x 150           &   140   &   3750       \\ \bottomrule
\end{tabular}
}
\vspace{-1.5em}
\end{table}

\subsection{Autonomous Docking Simulation Experiments}
\label{sec:proposed-method}
\begin{figure*}[!htb]
\centering
\subfigure[]{
  \includegraphics[width=0.23\linewidth]{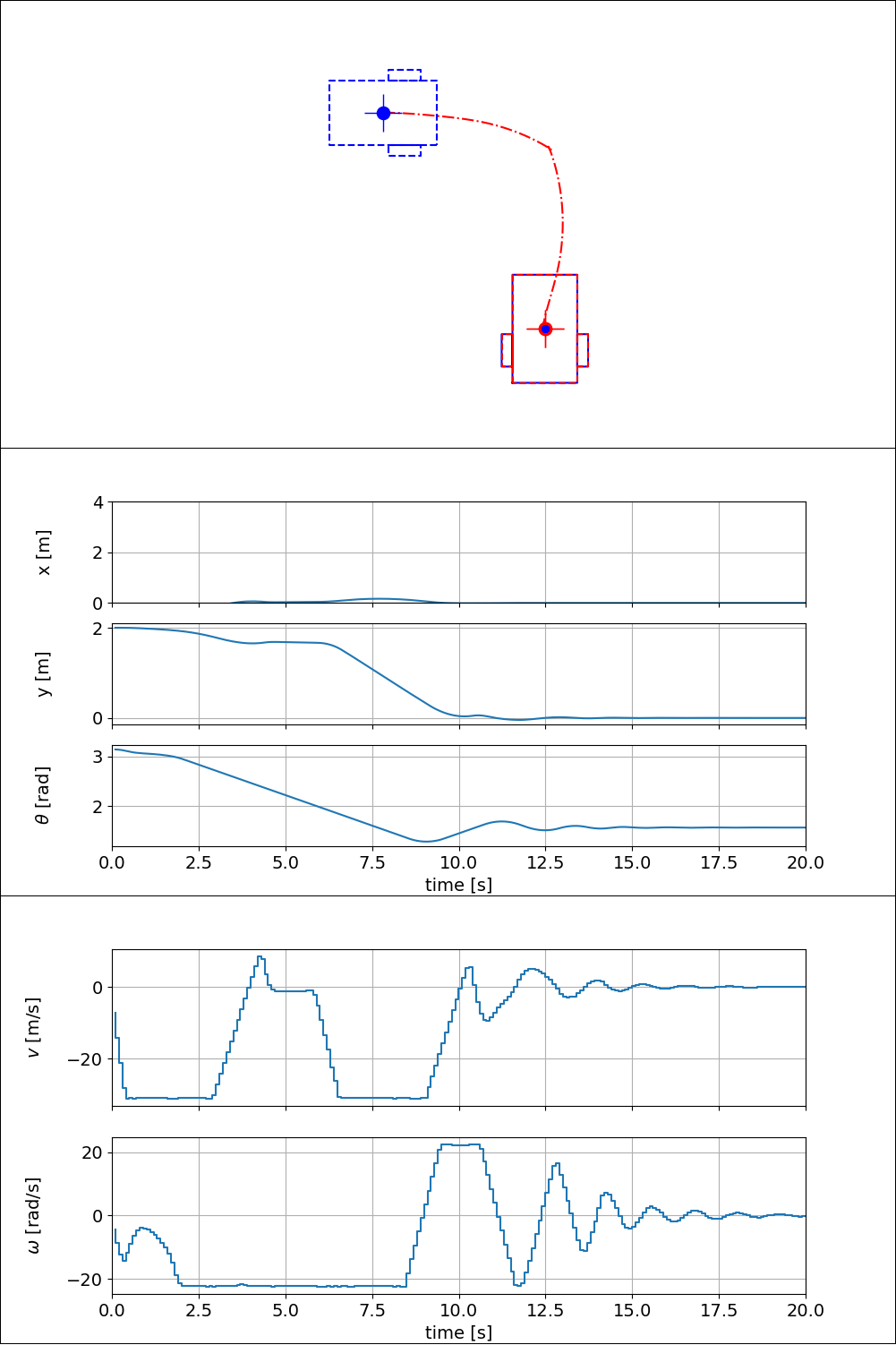}
  }
  \subfigure[]{
  \includegraphics[width=0.23\linewidth]{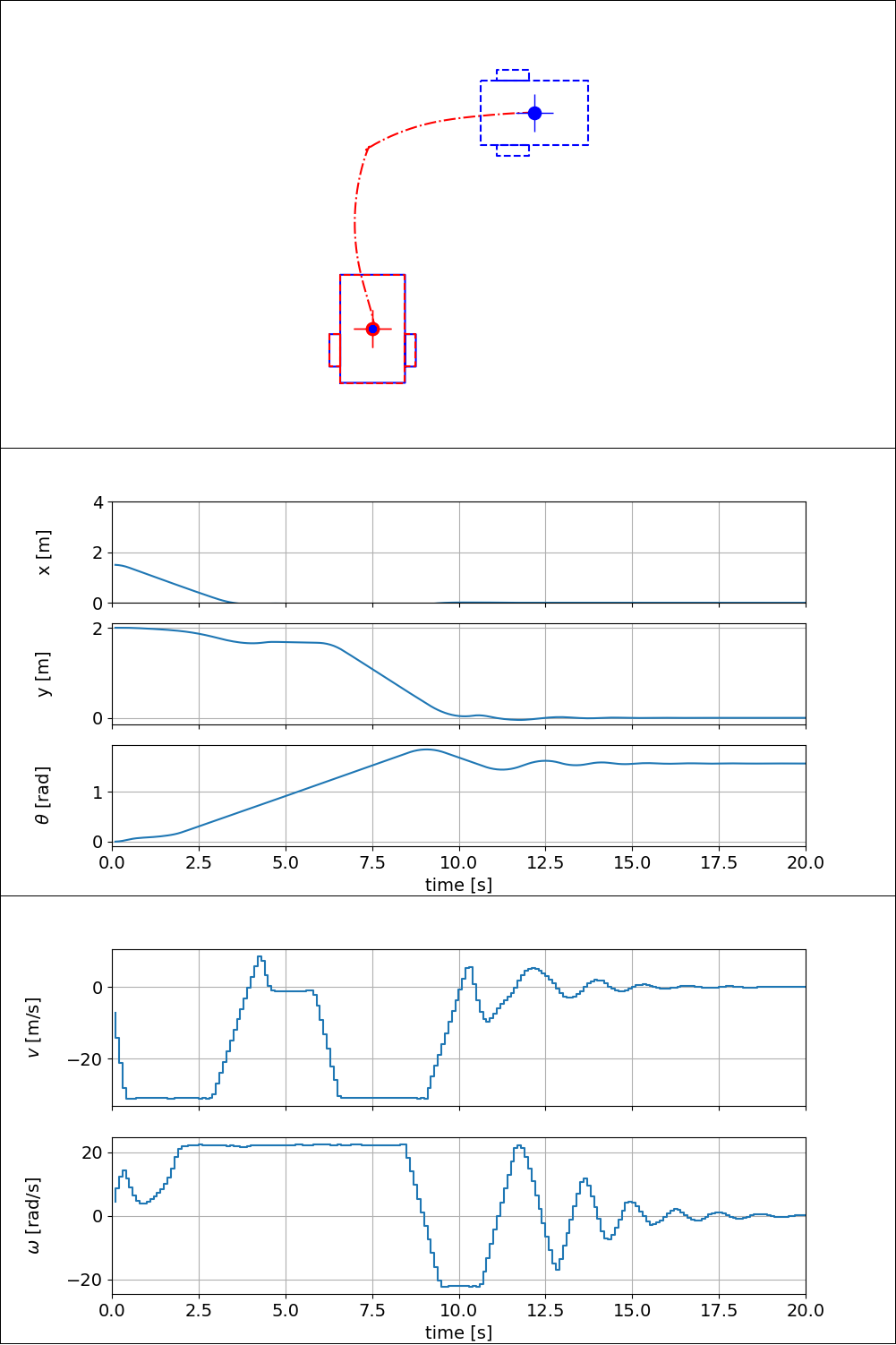}
  }
  \subfigure[]{
  \includegraphics[width=0.23\linewidth]{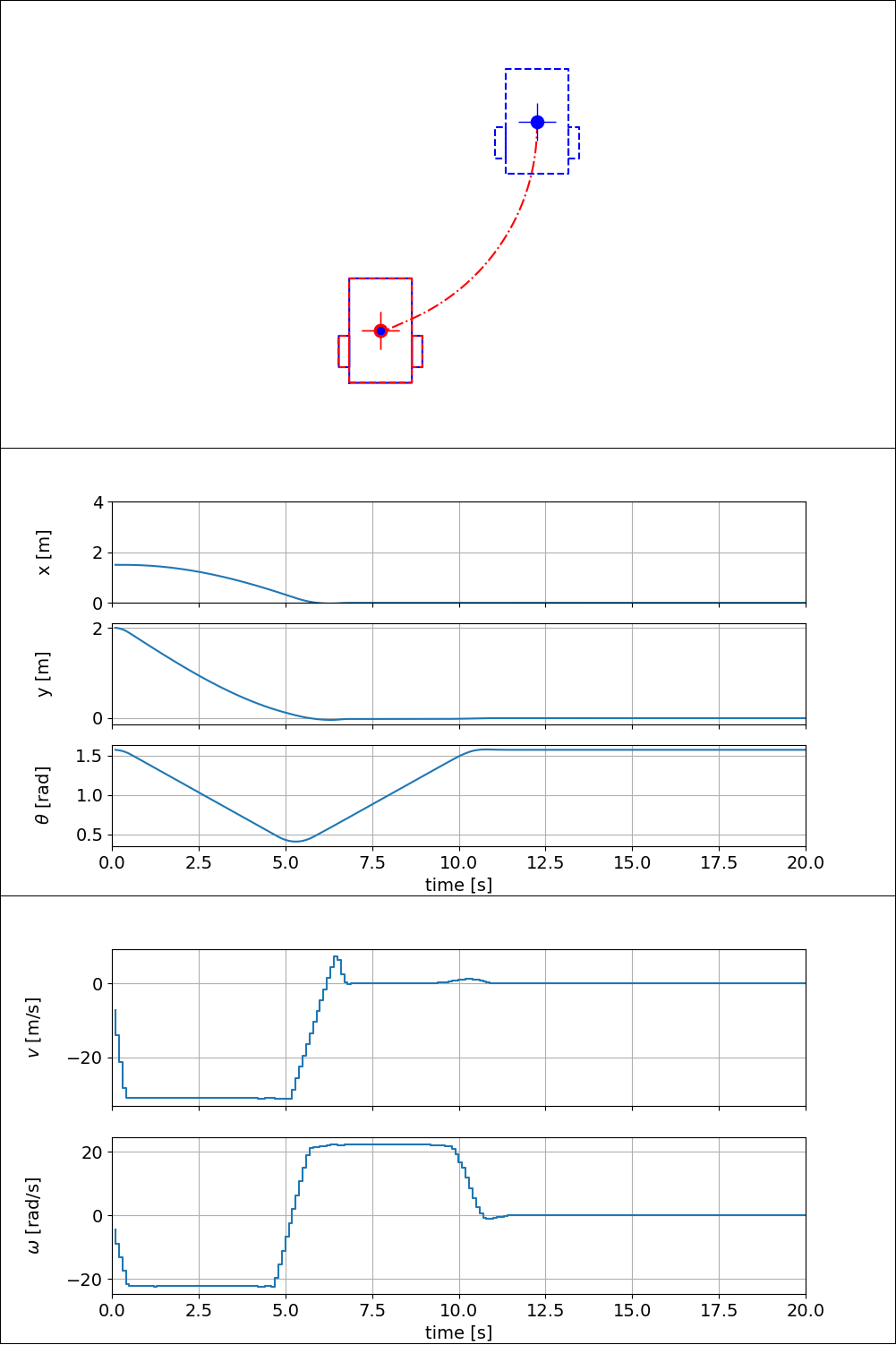}
  }
  \subfigure[]{
  \includegraphics[width=0.23\linewidth]{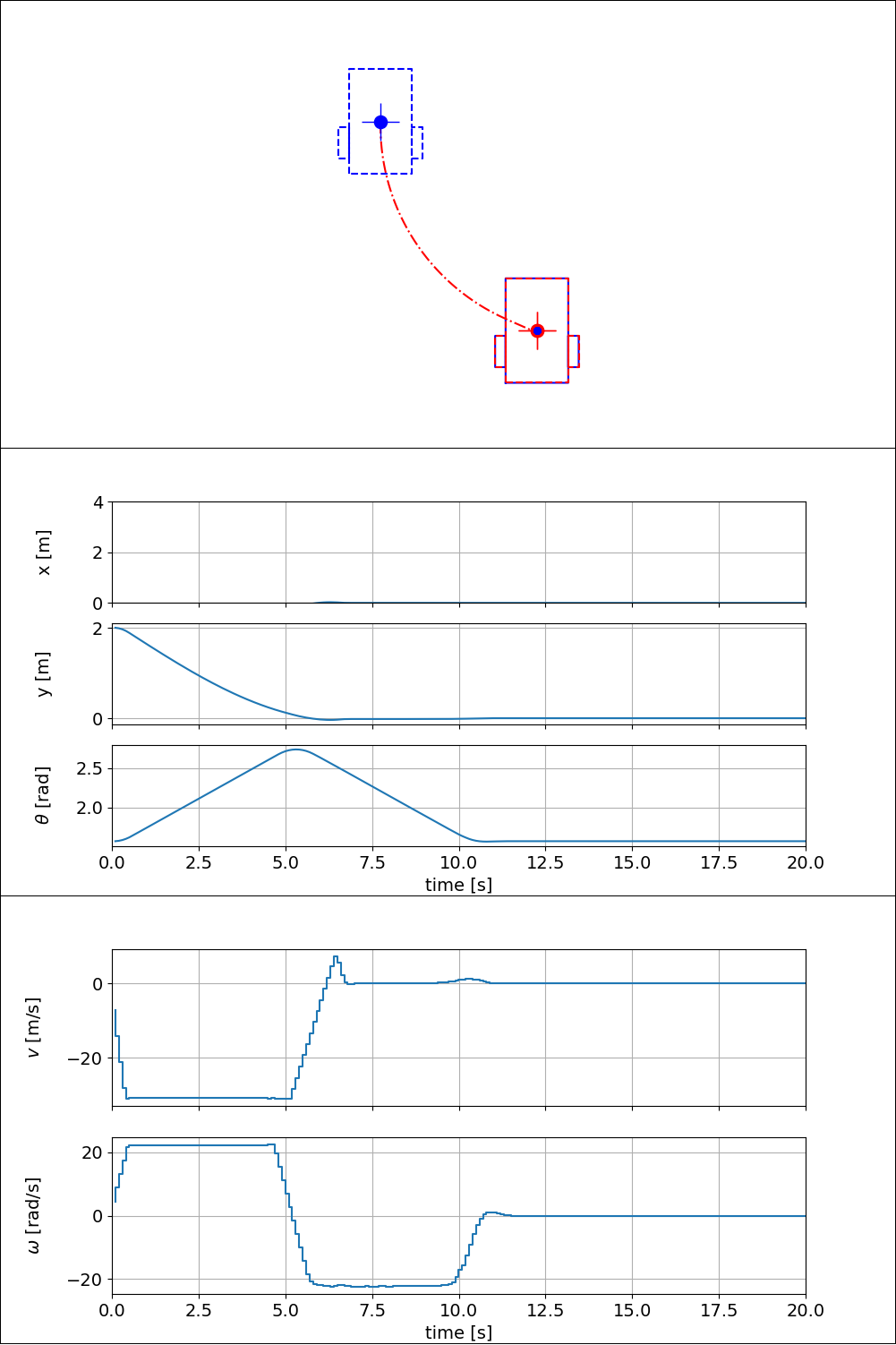}
  }
  \vspace{-0.5em}
\caption{Simulation experiment results of autonomous docking via non-linear model predictive control in four different scenarios.}
\vspace{-0.5cm}
\label{fig:docking-simulation-results}
\end{figure*}

Fig.~\ref{fig:docking-simulation-results} shows the simulation experiment results of autonomous docking via non-linear model predictive control in four different scenarios. 
It can be found from the results that the proposed non-linear model predictive controller has been successfully achieve the desired controller design requirement to generate realtime trajectory for the vehicle from four different initial poses to achieve the desired docking position and docking heading angle within the deviation tolerances.
Table~\ref{tab:experiment-error} shows the the position and heading angle deviations from the simulation experiment results of autonomous docking.
It can be found from the results that the average position deviation is $0.002132$ meter and the the average heading angle deviation is $0.000381^{\circ}$.

\begin{table}[t]
\centering
\caption{The position and heading angle deviations from the simulation experiment results of autonomous docking.}
\vspace{-0.1cm}
\label{tab:experiment-error}
\resizebox{\linewidth}{!}{
\begin{tabular}{@{}ccc@{}}
\toprule
\parbox{0.3\linewidth}{
Initial Pose \\ (x [m], y [m], $\phi$ [$^{\circ}$])
}
& 
\parbox{0.3\linewidth}{
Euclidean Position \\Error (m) 
}
& 
\parbox{0.3\linewidth}{
Heading Angle \\Error ($^{\circ}$)
}
\\ \midrule
(-1.5, 2.0, 90$^{\circ}$)           &   0.00281839  &  0.00045076                     \\
(1.5, 2.0, -90$^{\circ}$)           &  0.00282463   &   0.00071455                    \\
(1.5, 2.0, 0$^{\circ}$)          &  0.00144737   &   0.00015409                    \\
(-1.5, 2.0, 0$^{\circ}$)           &   0.00143895   &   0.0002046       \\
\midrule
\textbf{Average}           &   0.002132   &   0.000381
\\\bottomrule
\end{tabular}
}
\vspace{-1.5em}
\end{table}

\section{Conclusion}
\label{sec:conclusion}
This paper presents a proposed method of autonomous control for docking tasks of a single-seat personal mobility vehicle.
We proposed a non-linear model predictive control (NMPC) based visual servoing to achieves the desired autonomous docking task.
The proposed method is designed to be implemented on a four-wheel electric wheelchair platform, with two independent rear driving wheels and two front castor wheels.
A series of simulation experiments has been systematically conducted to evaluate the performance of the proposed controller method to achieved the controller design requirement.
The experimental outcomes obtained from various scenarios in the simulations provide evidence that the proposed controller technique is capable of achieving the controller design requirements for accomplishing the autonomous docking task.
Further study on this issue will be focused on integrating the proposed controller into the physical platform and conducting rigorous experiments and evaluations in a real-world setting.

\section*{Acknowledgment}
This work is fully funded by the Research Organisation for Electronics and Informatics, National Research and Innovation Agency (Badan Riset dan Inovasi Nasional - BRIN).

\bibliographystyle{IEEEtran}
\bibliography{references}

\end{document}